\documentclass[sigconf]{acmart}
\usepackage[encapsulated]{CJK}
\usepackage{multirow}
\usepackage{afterpage}
\usepackage{enumitem}
\renewcommand\footnotetextcopyrightpermission[1]{}
\AtBeginDocument{%
  }

\setcopyright{acmlicensed}
\copyrightyear{2018}
\acmYear{2018}
\acmDOI{XXXXXXX.XXXXXXX}
\acmConference[Conference acronym 'XX]{Make sure to enter the correct
  conference title from your rights confirmation email}{June 03--05,
  2018}{Woodstock, NY}
\acmISBN{978-1-4503-XXXX-X/2018/06}

\begin{document}

\title{Learning Speaker-Invariant Visual Features for Lipreading}

\author{Yu Li, Feng Xue, Shujie Li, Jinrui Zhang, Shuang Yang, Dan Guo, Richang Hong}









\begin{abstract}

 Lipreading is a challenging cross-modal task that aims to convert visual lip movements into spoken text. Existing lipreading methods often extract visual features that include speaker-specific lip attributes (e.g., shape, color, texture), which introduce spurious correlations between vision and text. These correlations lead to suboptimal lipreading accuracy and restrict model generalization. To address this challenge, we introduce SIFLip, a speaker-invariant visual feature learning framework that disentangles speaker-specific attributes using two complementary disentanglement modules (\textit{Implicit Disentanglement} and \textit{Explicit Disentanglement}) to improve generalization. Specifically, since different speakers exhibit semantic consistency between lip movements and phonetic text when pronouncing the same words, our implicit disentanglement module leverages stable text embeddings as supervisory signals to learn common visual representations across speakers, implicitly decoupling speaker-specific features. Additionally, we design a speaker recognition sub-task within the main lipreading pipeline to filter speaker-specific features, then further explicitly disentangle these personalized visual features from the backbone network via gradient reversal. Experimental results demonstrate that SIFLip significantly enhances generalization performance across multiple public datasets. Experimental results demonstrate that SIFLip significantly improves generalization performance across multiple public datasets, outperforming state-of-the-art methods.

 
 
\end{abstract}





\maketitle
\section{Introduction}
Lipreading aims to accurately recognize spoken text by analyzing lip movements, regardless of the presence of audio \cite{petridis2018end}. This field has attracted significant attention due to its potential applications in areas such as silent translation and public safety \cite{xu2020discriminative}. In recent years, significant advancements in deep learning have greatly improved lipreading performance \cite{kim2023lip, bulzomi2023end}.


Lipreading methods can be divided into word-level \cite{prajwal2022sub, 10504606} and sentence-level \cite{kumar2019lipper, chung2017out, chen2020lipreading, adeel2019lip, miao2020part} categories. Word-level lipreading focuses on recognizing isolated words, while sentence-level lipreading aims to understand the semantics of continuous lip movement sequences. This study primarily focuses on the latter, which involves more complex modeling due to longer temporal dependencies and richer semantic information. Existing sentence-level lipreading methods typically adopt deep learning frameworks, wherein the front-end adopts deep neural networks to extract visual features, and the back-end decodes these features into text. Various approaches have been proposed to improve model performance. For example, LipNet \cite{assael2016lipnet} was the first end-to-end sentence-level lipreading method, enabling continuous sentence prediction and laying the foundation for subsequent research. Zhang \textit{et.al} \cite{zhang2024es3} proposed learning audio-visual speech representations by modeling cross-modal shared, unique, and synergistic information. However, these methods often overlook the issue of generalization. As highlighted and empirically validated in \cite{10.1007/978-3-031-20059-5_33}, the features learned by lipreading models often contain speaker-specific information, which significantly hinders their ability to generalize to \textit{unseen speakers (speakers who are not present in the training set) }\cite{luo2023learning}. This challenge forms the primary motivation of our work. The root of this problem lies in the considerable variations in lip shape/color across speakers, as shown in Figure 1. Consequently, existing methods tend to extract visual features that inadvertently encode speaker-specific attributes, leading to spurious correlations between visual features and text. This impairs the model to accurately capture common visual features essential for speech content, i.e., speaker-invariant visual features, which in turn impacts its generalization ability. To mitigate speaker variability, Kim \cite{10.1007/978-3-031-20059-5_33} proposed a speaker-adaptive lipreading method based on user-dependent padding, which incorporates speaker-specific cues into the padding regions of convolutional layers without modifying the pretrained weights. While this method provides a lightweight adaptation mechanism, it still depends on injecting speaker-specific information. To address this fundamental limitation, our work focuses on decoupling speaker-specific visual features to learn more robust speaker-invariant representations across diverse speakers.

\begin{figure}
\centering
\includegraphics[width=\columnwidth]{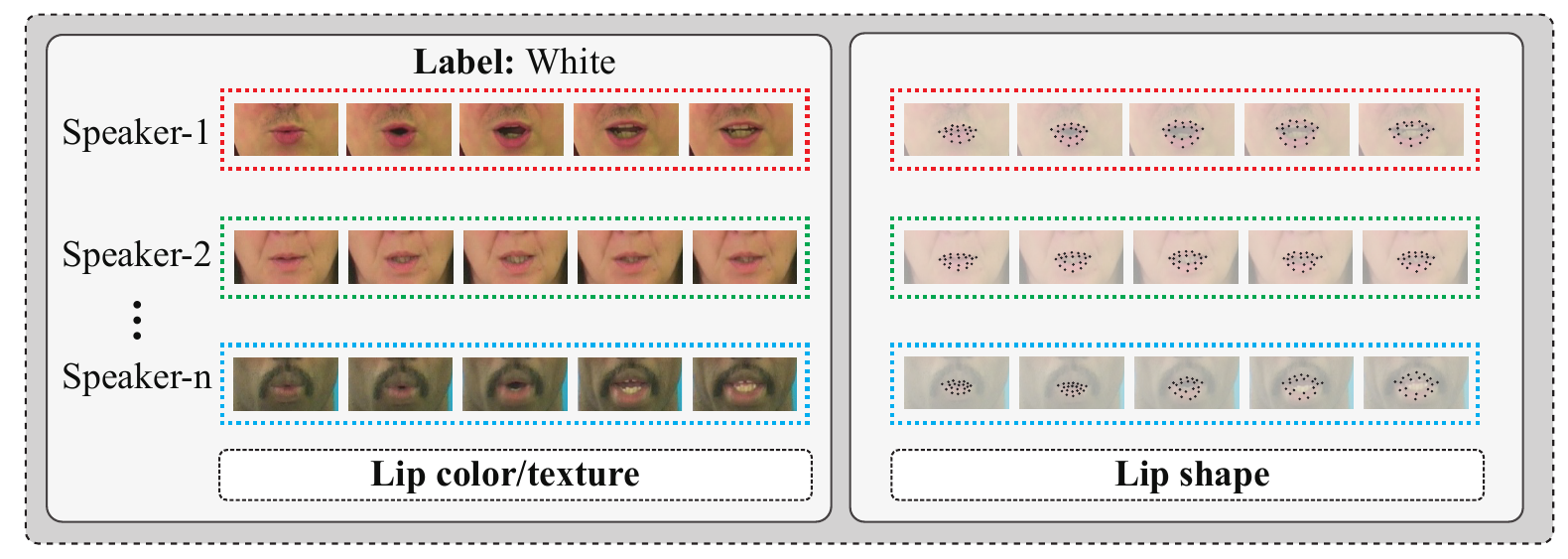}
\caption{Different speakers exhibit significant visual variations in lip color/shape when pronouncing the same word \textit{'white'}. These speaker-specific visual features limit the generalization of the model.}
\label{fig:motivation}
\end{figure}

Learning clean and sufficiently informative lip visual features is essential for improving the generalization of lipreading models. In \cite{10.1007/978-3-031-20059-5_33}, research shows that emphasizing speaker-invariant visual features during encoding and capturing the common features of different speakers when pronouncing the same text can effectively  improve generalization. The key to learning speaker-invariant lip visual features lies in effectively filtering out and disentangling speaker-specific features during network training while preserving recognition accuracy. To this end, in this paper, we propose a novel lipreading framework that learns speaker-invariant visual features by exploring both implicit and explicit disentanglement to decouple speaker-specific features in the backbone network.

In the speaker-specific features \textit{implicit disentanglement} module, we extract speaker-invariant visual feature by aligning spoken text with the visual features of different speakers, thereby improving lipreading accuracy. Since learning speaker-invariant visual features essentially involves filtering out speaker-specific visual (lip color/shape), this can be considered as a form of \textit{implicit disentanglement} of speaker-specific features. Considering the one-to-many mapping between the same spoken text and the visual features of different speakers, inspired by the cross-modal alignment strategy in CLIP \cite{radford2021learning}, we employ cross-modal contrastive learning to align the visual features of different speakers towards a fixed textual vector space, with the goal of learning speaker-invariant features. In lipreading, the subtle temporal variations in a speaker's lip movements (e.g., tongue positioning, lip aperture) evolve continuously across frames. Each frame's visual information carries critical features for distinguishing phonemes or word boundaries, directly impacting semantic understanding. Existing video pre-trained models (e.g., Video CLIP) \cite{fang2021clip2video, cheng2023cico} rely on coarse-grained cross-modal contrastive learning (e.g., global video-text alignment), which works for action classification but fails to achieve the frame-wise alignment essential for lipreading. To address this, we propose a frame-level label cross-modal alignment strategy, which forces the model to establish precise visual-semantic mappings by assigning corresponding text labels to each frame. This fine-grained supervision mechanism compels the model to learn high-quality text-aligned representations, significantly enhancing decoding capabilities for continuous lip movements while improving both recognition accuracy and generalization.
We further design an \textit{explicit disentanglement} module to eliminate potential residual speaker-specific features that may remain after the implicit disentanglement process. The core idea is to first filter out speaker-specific features and then disentangle them from the backbone network. To achieve this, we first design a speaker identification sub-task within the lipreading framework to filter out speaker-specific features. Specifically, the visual features extracted by the visual encoder in the backbone network are used for speaker classification, which guides the model to identify speaker-specific features that are strongly associated with speaker identity (e.g., lip shape). Subsequently, we incorporate a gradient reversal layer to invert the gradient of the speaker identification loss, forcing the visual encoder in the backbone to suppress speaker-specific features when updating its parameters. This explicit disentanglement process further improves the generalization of visual representations across different speakers.

The main contributions of this paper are summarized below:

1) We highlight the importance of disentangling speaker-specific features and learning speaker-invariant visual features to improve the generalization of lipreading models. 

2) We propose a framework SIFLip that learns speaker-invariant visual features for lipreading, which disentangles speaker-specific features by designing both explicit and implicit disentanglement modules, enabling the model to learn common features associated with spoken text, improving the generalization of the lipreading model to unseen speakers.

3) We conducted extensive experiments on two benchmark lipreading datasets to validate the effectiveness of the proposed SIFLip model. 

\section{Related Work}
In this section, we first briefly review deep learning-based lipreading methods. We then introduce feature disentanglement techniques and related methods for cross-modal learning.
\subsection{\textbf{Deep Learning-based Lipreading Methods}}
Lipreading, which converts video to text by analyzing lip movements, has advanced significantly with deep learning frameworks \cite{ren2021learning, ma2021towards, martinez2020lipreading, peymanfard2022lip}. Existing lipreading methods primarily focus on modifying neural network architectures to improve accuracy. For example, LipNet \cite{assael2016lipnet} is the first end-to-end lipreading model capable of continuous sentence recognition. To better capture global semantics across video frames, attention mechanisms have been incorporated into lipreading models \cite{xu2018lcanet, xue2023lcsnet, son2017lip, zhao2019cascade}, improving the ability to track subtle lip movements. Transformer-based architectures \cite{10447378, 10023442, 10141862, 9897760} further advance lipreading by enabling parallel training and faster convergence compared to RNN, with LipCH-Net \cite{Zhang_Gong_Dai_Yang_Liu_Liu_2019} being the first Transformer-based Chinese lipreading model.

The existing deep learning-based methods have achieved good recognition performance, but their generalization still has certain limitations, especially when it comes to speakers who have not appeared in the training set, the error rate is relatively high. In this paper, we propose learning speaker-invariant visual features to improve model generalization.

\subsection{\textbf{Feature Disentanglement}}
 Feature disentanglement \cite{xu2021variational, niu2020video, wang2022pose} improves generalization by separating domain-invariant (e.g., speaker-invariant features) and domain-specific (e.g., speaker-specific features) components. For example, in object detection, Liu \textit{et al.} \cite{liu2022decompose} applied feature disentanglement at global and local levels to remove source domain-specific information. Similarly, Deng \textit{et al.} \cite{deng2021informative} proposed an informative feature disentanglement module to separate high-level semantic features from both source and target domains, improving alignment in unsupervised domain adaptation. In face representation, Bortolato \cite{bortolato2020learning} proposed a feature disentanglement autoencoder PFRNet to suppress sensitive attributes, such as gender. 

For lipreading, speaker-specific features hinder model generalization. We improve generalization by disentangling speaker-specific visual features to learn speaker-invariant representations.

 \begin{figure*}[hbt]
\centering
\includegraphics[width= \textwidth]{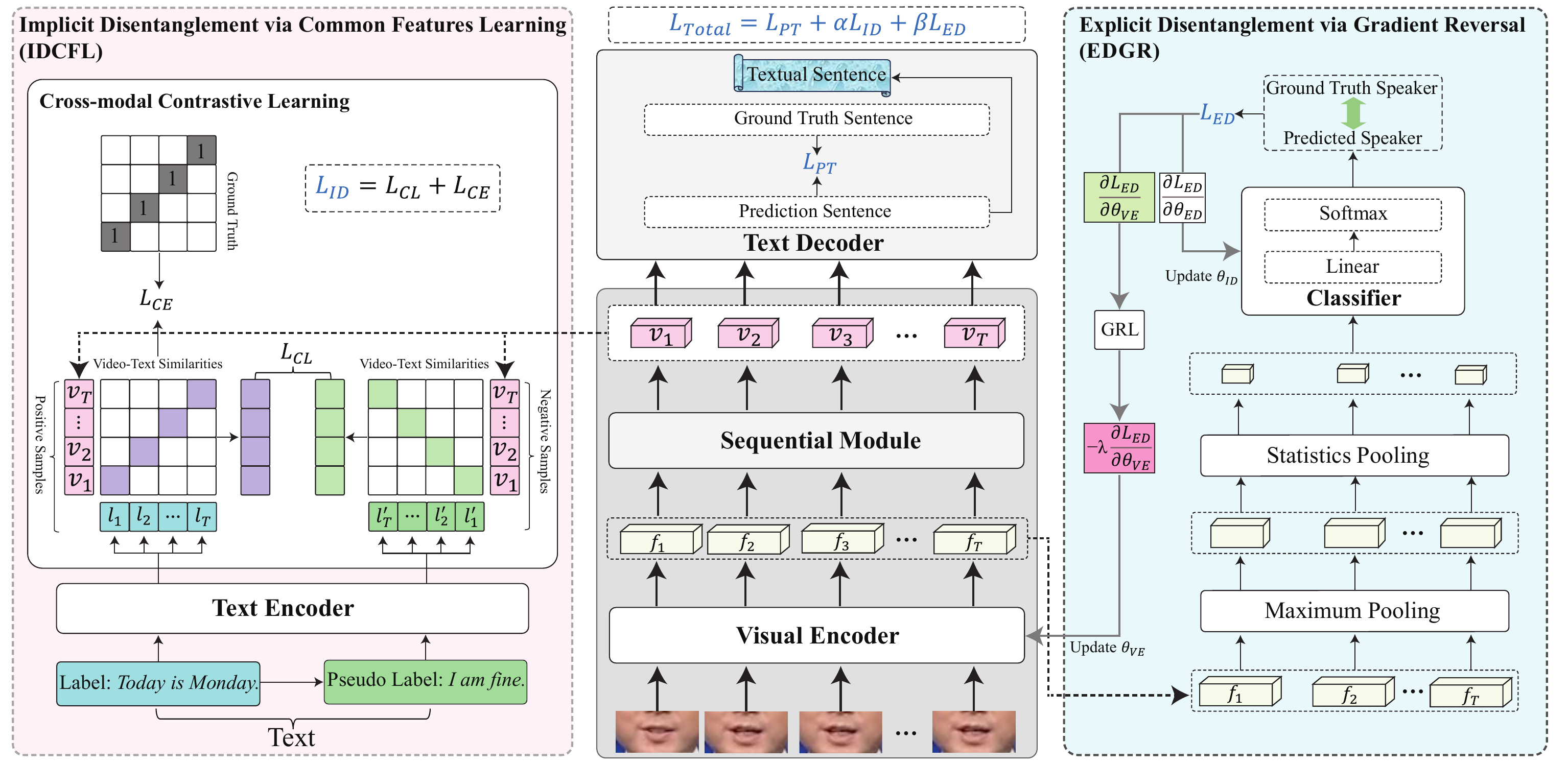}
\setlength{\abovecaptionskip}{0.05cm}  
\caption{Overview of the proposed SIFLip framework. SIFLip learns speaker-invariant visual features by disentangling speaker-specific features. IDCFL achieves implicit disentanglement by aligning cross-modal features through the calculation of similarity between visual and text features. EDGR achieves explicit disentanglement by reversing the gradient of speaker classification, wherein the classification is based on speaker-specific features filtered through pooling operations. Finally, the visual features are fed into a seq2seq model to be decoded into text.}
\label{fig:model}
\end{figure*}
\subsection{\textbf{Cross-Modal Learning}}
To enhance cross-modal understanding between visual and textual modalities, recent works \cite{li2023blip, radford2023robust, wang2023exploring, luo2022clip4clip} have proposed using text as a supervisory signal to improve the representation of visual features. For instance, Radford \textit{et al.} \cite{radford2021learning} proposed CLIP, which jointly pre-trains an image and text encoder by predicting correct image-text pairs. Similarly, Fang \textit{et al.} \cite{fang2021clip2video} introduced CLIP2Video, which aligns video and text in a shared embedding space to facilitate video-language understanding. In sign language recognition, Zhou \textit{et al.} \cite{zhou2023gloss} pre-trained a visual-language model with text supervision to bridge the semantic gap between visual and textual representations.

Lipreading is a cross-modal task that converts video into text. We use text as a stable supervisory signal to learn common features by cross-modal alignment, implicitly disentangling speaker-specific visual features for accurate video-to-text mapping.
\section{The Proposed Method}

The architecture of the proposed SIFLip is illustrated in Figure \ref{fig:model}, which comprising four modules: 1) a visual feature extraction module that extracts visual features of lip region; 2)an implicit disentanglement module(IDCFL) that learns common features aligned with textual semantics to implicitly disentanglement speaker-specific features;3)an explicit disentanglement module(EDGR) that explicitly filter speaker-specific features and decouples them from the backbone network; 4) a text decoding component that decoding visual features to text.

\subsection{Visual Features Extractor}
\noindent\textbf{Visual Features Embedding.}
Given a video clip $X\in \mathbb{R}^{T \times H \times W \times 3}$ with $T$ frames as input, we use 3D-CNN to extract short-term spatiotemporal features of lip movements. After each convolutional layer, ReLU and max pooling are applied. To mitigate overfitting, dropout layers are also included.

\noindent\textbf{Sequential Module.} 
The sequential module captures temporal dependencies in visual features $f = \{f_1, \ldots, f_T\}$ extracted by the 3D-CNN. We first use a bidirectional GRU to capture both forward and backward temporal information from the visual features, modeling the dynamic features of lip movements: 
\begin{equation}
f_i = [\overrightarrow{f_i}, \overleftarrow{f_i}].
\end{equation}
To facilitate information exchange, we further append a 6-layer Transformer after the 3D-CNN and Bi-GRU to model the long-term temporal dependencies of the visual features:
\begin{equation}
v_i = \text{Encoder}(f_i + \text{PE}_{1:T}) \in \mathbb{R}^{T \times d},
\end{equation}
where PE denotes Position Encoding, $d$ is the dimensionality of the visual embedding vectors.


\subsection{\textbf{Implicit Disentanglement via Common Features Learning (IDCFL)}}
 Due to the inherent semantic consistency between lip movements and their corresponding text when different speakers pronounce the same words, we design a cross-modal consistency learning module to obtain speaker-independent visual feature representations of lip movements. Specifically, within the multi-modal embedding space, we enforce visual-text alignment and use text-stabilized embeddings as supervisory signals to learn common visual features across speakers, thereby implicitly disentangling and eliminating speaker-specific visual attributes.
 
 To achieve fine-grained semantic alignment between lipreading videos and text, we propose a frame-level label cross-modal alignment method that assigns corresponding text labels to each video frame, establishing a frame-by-frame semantic mapping relationship. Specifically, given a video-text pair \(\langle X, t \rangle\), we construct a positive pair \(\langle X, L \rangle\) and a negative sample pair \(\langle X, L' \rangle\). By learning cross-modal alignment in a multimodal joint embedding space, the model effectively achieves accurate video-text matching. Since text features remain stable regardless of speaker differences, text serves as a reliable supervisory signal to guide the model in learning common visual features that align with the semantics of the text. During this process, the model implicitly disentangles speaker-specific features, facilitating the extraction of common features.

\begin{figure}[!t]
\centering
\includegraphics[width=0.95\columnwidth]{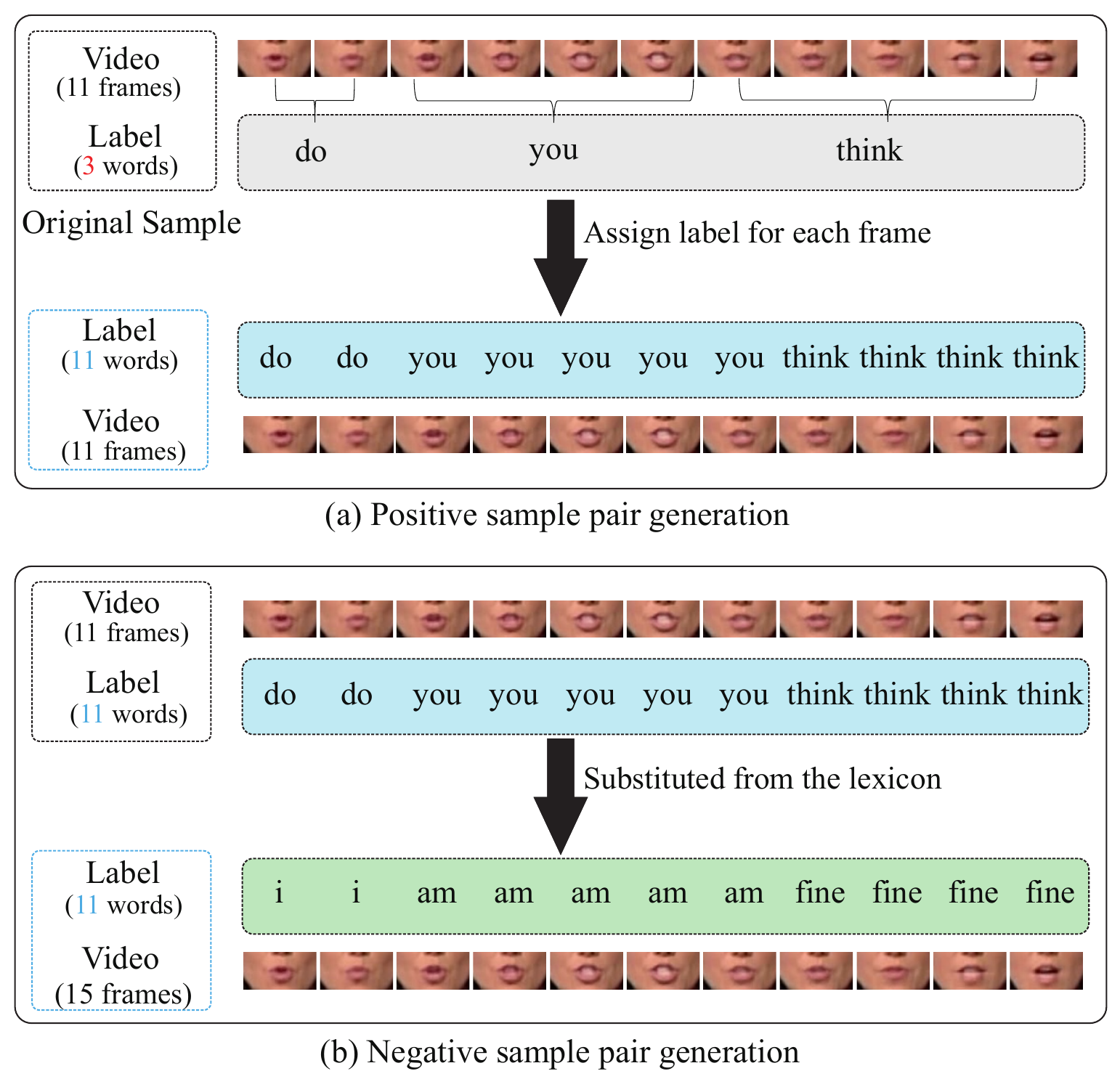}
\caption{Illustration of positive and negative sample pair generation. Each frame is assigned a corresponding label based on word boundaries in the video frames, forming a positive video-text pair. The negative sample pair is generated by replacing all labels in the positive sample pair.}
\label{fig:Assign}
\end{figure}


\noindent \textbf{Positive and Negative Sample Pair Generation.} Given a video-text pair \( (X, t) \), where \( X = \{x_1, x_2, \cdots, x_T\} \) denotes a video clip, and \( t = \{t_1, t_2, \cdots, t_n\} \) represents the corresponding text sequence of length \( n \) (\( n < T \)), we first use MFA to determine word boundaries within the video. Each frame is then assigned a label based on these boundaries, yielding a label sequence \( L = \{l_1, l_2, \cdots, l_T\} \) with length \( T \). This process ensures that the video and text sequences are aligned at the frame level, forming a positive sample pair \( (X, L) \). As shown in Figure \ref{fig:Assign} (a), the word boundaries segment the video into different regions, and each frame is assigned a corresponding label from the text sequence, enabling fine-grained supervision for model training. 

To create a negative sample, we replace all characters in the labels of the positive sample with alternatives randomly selected from the lexicon, generating pseudo-text sequences. The video clip is then paired with these pseudo-text sequences to form the negative sample pair \( (X, L') \), as shown in Figure \ref{fig:Assign} (b). This replacement results in a mismatch between the text sequence and the video, making the negative sample semantically different from the positive sample.

\noindent\textbf{Text Encoding.} We use a Transformer Encoder to extract text features. The text sequence \( L = \{l_1, l_2, \cdots, l_T\} \) is first mapped to a learnable embedding matrix \( Q \in \mathbb{R}^{T \times d} \), with sinusoidal positional encodings added. The combined representations are then passed through an \( N \)-layer Transformer Encoder to model the temporal dynamics of the sequence:
 \begin{equation}
Q = \text{Embedding}(l_i) \in \mathbb{R}^{T \times d},
\end{equation}
\begin{equation}
l_{i} = \text{Encoder}(Q + \text{PE}_{1:T}) \in \mathbb{R}^{T \times d}.
\end{equation}


\noindent\textbf{Cross-Modal Similarity.} Inspired by CLIP, we first calculate the similarity for the positive sample pair \(\left \{ \left (v_{i}, l_{i}^{pos}  \right )  \right \} _{i=1}^{T}\) and negative sample pair \(\left \{ \left (v_{i}, l_{i}^{neg}  \right )  \right \} _{i=1}^{T}\). This results in two similarity matrices \( \textbf{E}_{(v,l)}^{{\rm pos}}\in \mathbb{R}^{T \times T}\) and \( \textbf{E}_{(v,l)}^{{\rm neg}} \in \mathbb{R}^{T \times T} \). Each element in \( \mathbf{E}_{(v,l)} \) represents the similarity between the visual feature \( v_i \) and the text feature \( l_i \).

\noindent\textbf{Cross-modal Contrastive Learning.} After obtaining the similarity matrices for both the positive and negative sample pair, we perform contrastive learning between \( \textbf{E}_{(v,l)}^{{\rm pos}} \) and \( \textbf{E}_{(v,l)}^{{\rm neg}} \), using the InfoNCE loss function to bring together the embeddings of matching image-text pair while separating the embeddings of mismatched image-text pair. This can be formalized as:

\[
\mathcal{L}_{CL} = -\frac{1}{T} \sum_{i=1}^{T} \log \frac{\exp\left(e_{i}^{{\rm pos}} / \tau\right)}{\exp\left(e_{i}^{{\rm pos}} / \tau\right) + \sum_{j=1}^{T} \exp\left(e_{j}^{{\rm neg}} / \tau\right)},
\]
where \( e_{i}^{{\rm pos}} \) and \( e_{i}^{{\rm neg}} \) are the diagonal elements of \( \textbf{E}_{(v,l)}^{{\rm pos}} \) and \( \textbf{E}_{(v,l)}^{{\rm neg}} \), respectively, while \( \tau \) is a trainable temperature parameter.

To further improve the alignment between visual features and text, we use the cross-entropy loss function to compute the loss between the similarity matrix of the positive sample pair \( \textbf{E}_{(v,l)}^{{\rm pos}} \) and the ground truth matrix \( \textbf{E}_{(v,l)}^{{\rm G}} \):
\begin{equation}
\mathcal{L}_{CE} = - \sum \textbf{E}_{(v,l)}^{{\rm G}} \log( \textbf{E}_{(v,l)}^{{\rm pos}})
\end{equation}

where $\textbf{E}_{(v,l)}^{{\rm G}}$ is a $T×T$ identity matrix, with ones on the diagonal and zeros in the off-diagonal elements.

We define the above two losses as $\mathcal{L}_{ID}$, as follows:
\begin{equation}
\mathcal{L}_{ID} = \mathcal{L}_{CL}+ \mathcal{L}_{CE}.
\label{eq:ID}
\end{equation}


\subsection{\textbf{Explicit Disentanglement via Gradient Reversal (EDGR)}}

In addition to the implicit disentanglement module described in Section 3.2, to further disentangle speaker-specific features from the backbone network, we designed an explicit disentanglement module for speaker-specific visual lip features. Specifically, we designed a speaker recognition sub-task within the main lipreading pipeline to filter out speaker-specific visual features. Then, we further explicitly disentangled these personalized visual features from the backbone network via gradient reversal, thereby improving the generalization ability of visual representations across speakers.

To extract speaker-specific visual features, the intermediate feature representation \( f_i \) undergoes a series of pooling operations. Initially, max pooling is applied to compress the spatial dimensions, highlighting salient speaker-specific features and generating the intermediate feature \( f_{i}^{'} \). Subsequently, statistical pooling is performed by concatenating the mean and standard deviation of \( f_{i}^{'} \) across the temporal dimension, forming a global speaker representation \( f_{i}^{''} \). This representation is then input into a classifier that computes the prediction probability \(y^{'} \) for each speaker. The module is trained with loss $\mathcal{L}_{ED}$, which is a cross-entropy loss:


\begin{equation}
y^{'} = {\rm Softmax}({\rm Linear}({\rm ReLU}({\rm Linear}(f_{i}^{''}))))
\end{equation}
\begin{equation}
 \mathcal{L}_{ED} =-{\textstyle \sum_{i=1}^{N}} y_i \log(y^{'}),
\end{equation}
where \(y^{'} \in \mathbb{R}^{N} \), \( N \) is the total number of speakers, and \( y_i \) denotes the ground truth speaker ID.

To mitigate the influence of speaker-specific features during training, we incorporate a Gradient Reversal Layer (GRL) \cite{ganin2016domain} into the framework based on feature disentanglement principles. Positioned between the feature extractor and speaker classifier, the GRL reverses the gradient of the speaker classification loss, guiding the model to explicit disentanglement speaker-specific features. Formally, let \(\mathcal{L}_{ED} \) denote the speaker classification loss. The visual encoder network parameters are denoted as \( \theta_{VE} \), and the speaker classifier parameters as \( \theta_{ED} \). The overall training is expressed as:
\begin{equation}
\theta_{ED} \leftarrow \theta_{ED} - \eta \frac{\partial L_{ED}}{\partial \theta_{ED}},
\end{equation}
\begin{equation}
\theta_{VE} \leftarrow \theta_{VE} - (-\lambda) \eta \frac{\partial L_{ED}}{\partial \theta_{VE}},
\end{equation}
where  \( \eta \) is the learning rate, $\lambda$ is the parameter.



As the speaker classifier accurately predicts the speaker's ID, the gradient reversal layer forces the model to penalize speaker-specific features, driving the backbone network to explicitly disentangle them.



\subsection{\textbf{Text Decoding via Seq2Seq Model}}

We adopt a sequence-to-sequence (seq2seq) model \cite{sutskever2014sequence} to decode visual features into text. The encoder processes the input feature sequence \( v = \{v_1, \allowbreak \ldots, \allowbreak v_T\} \), generating hidden states \( h_e^v \) , while the last hidden state initializes the decoder’s GRU. At each time step $j$, the decoder updates its hidden state \( h_d^{t} \) based on the previous prediction \( w^{t}_{j-1} \) and the prior hidden state: 

\begin{equation}
(h^{v}_{e})_{j}=GRU^{v}_{e}((h^{v}_{e})_{j-1},e^{v}), 
\end{equation}
\begin{equation}
(h^{t}_{d})_{j}=GRU^{t}_{d}((h^{t}_{d})_{j-1},e^{t}_{j-1}).
\end{equation}
The encoder output is attended at each time step to compute the context vector $(c^{v}_{t})_{j}$, which is used to generate the output $w$. The model is trained with predicted text loss \( \mathcal{L}_{PT} \):
\begin{equation}
(c^{v}_{t})_{j}=h^{v}_{e}\cdot att((h^{t}_{d})_{j},h^{v}_{e})), 
\end{equation}
\begin{equation}
P(w)= {\rm Softmax}({\rm MLP}((h^{t}_{d})_{j},(c^{v}_{t})_{j}),
\end{equation}
where $v$ and $t$ denote the visual and text modalities, respectively.

\begin{figure}[h]
\centering
\includegraphics[width=\columnwidth]{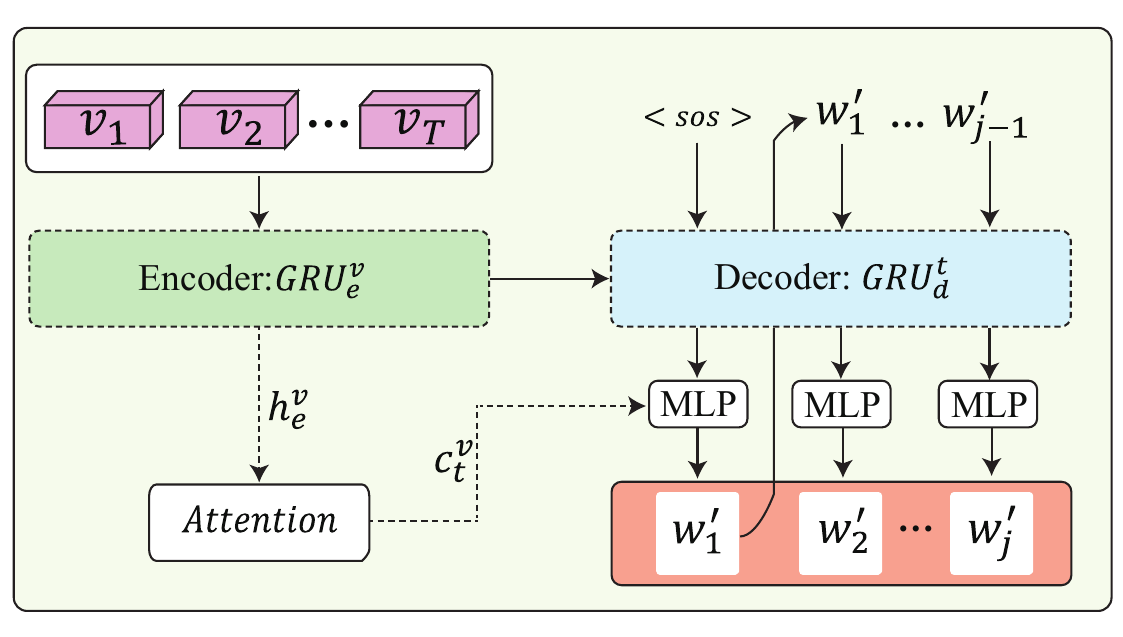}
\setlength{\abovecaptionskip}{0.1cm} 
\caption{Illustration of Seq2Seq module. The Seq2Seq module with an encoder-decoder structure that decodes visual features into text.}
\label{fig:seq2seq}
\end{figure}

\begin{equation}
\mathcal{L}_{PT} =-\frac{1}{n} {\textstyle \sum_{i=1}^{n}} {\textstyle \sum_{j=1}^{m}}w_{ij}log w_{ij}^{'},
\end{equation}

where \( w \) denote the ground-truth, \( n \) is the sequence length, and \( m \) is the number of word categories.

In addition, our model can be applied to Chinese datasets. For Chinese dataset, we implement a cascaded seq2seq \cite{zhao2019cascade,xue2023lipformer} model that first predicts pinyin, which is then converted into Chinese characters. Given the limited phonetic forms in pinyin, this intermediate step enhances cross-modal decoding accuracy. 

\begin{table*}[hbt!]
    \fontsize{9}{9}\selectfont
    \caption{Performance comparison with state-of-the-art methods on CMLR and GRID. -: results not available. Best results are in boldface. Imp indicates the absolute improvement of SIFLip relative to the best-performing baseline.}
    \label{tab:SOTA-CMLR}
    \centering
    \begin{tabular}{lcccc}
        \toprule
        \multirow{2}*{\textbf{Methods}}  
        & \multicolumn{2}{c}{\textbf{CMLR}} & \multicolumn{2}{c}{\textbf{GRID}} \\
        \cmidrule(lr){2-3} \cmidrule(lr){4-5} 
        & {\textbf{Seen Speakers (CER $\downarrow$)}} & {\textbf{Unseen Speakers (CER $\downarrow$)}} & {\textbf{Seen Speakers (CER $\downarrow$)}} & {\textbf{Unseen Speakers (CER $\downarrow$)}} \\
        \midrule
        LipNet\cite{assael2016lipnet} & 33.41 & 52.18  &4.8 & 17.5 \\ 
        WAS \cite{son2017lip}& 38.93 & -& 3.0 & 14.6  \\
        CSSMCM \cite{zhao2019cascade}  & 32.48 & 50.08& - & -  \\
        LIBS  \cite{2019Hearing} & 31.27 & - & - & -\\
        CALLip\cite{huang2021callip} &  31.18 & -  & 2.48 & - \\
        LCSNet \cite{xue2023lcsnet} & 30.03 & 46.98 & 2.3 & 11.6 \\
        Ai \textit{et.al} \cite{10141862}& - & - & 1.1 & - \\
        LipFormer \cite{xue2023lipformer} & 27.79 & 43.18& 1.45 & 9.64  \\
        \midrule
        \textbf{SIFLip} &  \textbf{20.55} &  \textbf{32.16} & \textbf{0.79} & \textbf{6.23} \\
        \midrule
        \textit{Imp (absolute)} & \textit{+7.24} & \textit{+11.02} & \textit{+0.31} & \textit{+3.41}  \\
        \bottomrule     
    \end{tabular}
\end{table*}


\subsection{\textbf{Model Optimization}}
To optimize SIFLip, we jointly minimize three loss functions: \( \mathcal{L}_{\text{PT}} \), \( \mathcal{L}_{\text{ED}} \), and \( \mathcal{L}_{ID} \). The predicted text loss \( \mathcal{L}_{\text{PT}} \) ensures that the text sequence predicted by the backbone network closely aligns with the ground truth. \( \mathcal{L}_{\text{ED}} \) facilitates the semantic alignment of visual features extracted by the backbone with text features, implicitly disentangling speaker-specific features. Meanwhile, \( \mathcal{L}_{ID} \) guides the branch to filter speaker-specific features, further assisting the backbone in explicitly disentangling them. The total loss is formulated as a weighted sum of these three losses: 

\begin{equation}
\mathcal{L}_{Total}  = \mathcal{L}_{PT} + \alpha  \mathcal{L}_{ID}+ \beta \mathcal{L}_{ED},
\label{eq:Total}
\end{equation}
where $\alpha$ and $\beta$ are hyperparameters.

\section{Experiments}

\subsection{Experimental Settings}
\subsubsection{\textbf{Dataset}} 
We validate the proposed method on two benchmark datasets: CMLR \cite{zhao2019cascade} and GRID \cite{cooke2006audio}.

\textbf{CMLR} \cite{zhao2019cascade} is the largest publicly available Chinese lipreading dataset, includes 11 speakers and over 100,000 sentences. For both unseen and seen speakers, we follow the division method in \cite{xue2023lipformer}.

\textbf{GRID} \cite{cooke2006audio}is a widely used English lipreading dataset, consisting of 32,823 sentences spoken by 33 speakers. We use the same dataset division method for unseen and seen speakers as in \cite{assael2016lipnet}. 



For the sentence-level datasets CMLR and GRID,, Word Error Rate (WER) (\%) is used as the evaluation metric \cite{huang2021callip, 2019Hearing, huang2021callip}.

\subsubsection{\textbf{Baselines.}} To evaluate the effectiveness of our proposed SIFLip, we compared our method with several baselines: LipNet \cite{assael2016lipnet}, WAS \cite{son2017lip}, CSSMCM \cite{zhao2019cascade}, LIBS \cite{2019Hearing}, CALLip \cite{huang2021callip}, LCSNet \cite{xue2023lcsnet}, Ai \textit{et.al} \cite{10141862}, LipFormer \cite{xue2023lipformer}.

\subsubsection{\textbf{Implementation Details.}} For each video clip, we use the DLib face detector \cite{amodio2018automatic} to locate facial regions in each frame and extract a 160 × 80-pixel mouth-centered crop as input. The visual encoder comprises a 3-layer 3D-CNN, while the Transformer encoder consists of 6-layer, each with a hidden size of 512 and 8 attention heads.  In Eq.~\eqref{eq:Total}, we set $\alpha$ = 0.5, $\beta$ = 2.0 for CMLR and GRID datasets.

The word boundaries in Sec 3.2 are obtained using the MFA (Montreal Forced Aligner). Specifically, MFA aligns audio with text labels to determine the start and end boundaries of each word in the audio. Given audio-video synchronization, this alignment enables accurate localization of the word boundaries in the video. Notably, during training, both video and text were used as inputs, whereas during testing, only video was used.

\subsection{\textbf{Overall Comparison}}
In this subsection, we compare SIFLip with other typical baselines. LipNet is the basic pipeline for lipreading. However, its CTC loss failed to converge during training on CMLR. Therefore, we replaced CTC with Seq2Seq module. Beyond evaluating seen speakers, we also assessed the model on unseen speakers to measure its generalization.
\noindent \textbf{Results on CMLR dataset.} Table \ref{tab:SOTA-CMLR} compares the proposed SIFLip with baselines. The results show that SIFLip significant performance improvements, particularly for unseen speakers, demonstrating its superior generalization. Specifically, compared to LipFormer, SIFLip achieves absolute improvements of +7.24 for seen speakers and +11.02 for unseen speakers. Moreover, its advantages over other baselines are even more evident. This can be attributed to its ability to disentangle speaker-specific visual features, effectively minimizing the impact of speaker variations.

\noindent \textbf{Results on GRID dataset.} As shown in Table \ref{tab:SOTA-CMLR}, SIFLip outperforms all baseline methods, both for seen and unseen speakers. Specifically, LipNet is the earliest lipreading method, achieves a CER of 4.8\% for seen speakers, while SIFLip reduces it to 0.79\%. WAS and CALLip uses audio-visual bimodal input, SIFLip achieves better performance with visual unimodal input. These improvements are due to SIFLip to learn speaker-invariant visual features.

\subsection{\textbf{Ablation Study}}
In the overall comparison experiment in Section 4.2, four datasets were used. Since the CMLR dataset is the most representative (it contains speaker identity information, and the spoken texts exhibit a certain degree of diversity in terms of length and content). Therefore, in this section, we use the CMLR as the test dataset for the ablation study.

\subsubsection{\textbf{Effect of IDCFL and EDGR}}
We first conduct ablation studies to validate the effectiveness of the IDCFL and EDGR modules in SIFLip. The following three variants are designed:
\begin{itemize}
\item {\textit{Base}}: It is the backbone of SIFLip, consisting of 3D-CNN, GRU, Transformer, and Seq2Seq. 
\item {\textit{Base w/ IDCFL}}: It adds the IDCFL module to the Base. 
\item {\textit{Base w/ EDGR}}: It adds the EDGR module to the Base.

\end{itemize}

\begin{table}[H]
\fontsize{10}{10}\selectfont
\caption{Ablation study of IDCFL and EDGR.}
\label{tab:hy-ALL}
    \centering
    \begin{tabular}{lccc}
        \toprule
        \multirow{2}{*}{\textbf{Methods}} & \multicolumn{2}{c}{\textbf{CMLR}} \\
        \cmidrule(lr){2-3}
         & \textbf{Seen (CER $\downarrow$)} & \textbf{Unseen (CER $\downarrow$)} \\
        \midrule
        Base  & 28.52 & 47.86 \\ 
         Base \textit{w/} IDCFL & 21.49 & 33.13 \\
         Base \textit{w/} EDGR & 26.22 & 39.02 \\
         \hline
         \textbf{SIFLip }& \textbf{20.55} & \textbf{32.16} \\
        \bottomrule
    \end{tabular}
\end{table}

Table~\ref{tab:hy-ALL} summarizes the ablation experiments evaluating the contributions of the IDCFL and EDGR modules in SIFLip, leading to the following observations. 

Among all variants of SIFLip, the Base performs the worst, as visual variations result in the extracted features containing substantial speaker-specific features, which limits generalization. In contrast, the performance of Base w/ IDCFL improves significantly. This is attributed to the fact that IDCFL employs text as a supervisor signal, which strengthens the visual-text alignment and guide the model to learn common visual features that are semantically consistent with the text, achieving implicit disentanglement of speaker-specific features. Meanwhile, Base w/ EDGR also outperforms the Base, confirming the effectiveness of the EDGR module in explicitly disentangling speaker-specific features to improving generalization. Notably, Base w/IDCFL achieves better performance than Base w/EDGR, suggesting that learning speaker-independent features aligned with semantic content plays a more critical role in improving generalization for the lipreading task. Finally, SIFLip, which integrates both IDCFL and EDGR, achieves the best overall performance, highlighting the complementary effects of implicit and explicit disentanglement. 


\subsubsection{ \textbf{Effect of Joint Loss}} 
We then conducted ablation studies to evaluate the contribution of each disentanglement within the IDCFL and EDGR modules. To achieve implicit disentanglement of speaker-specific features, the IDCFL module performs cross-modal alignment between visual and textual features by incorporating contrastive loss $\mathcal{L}_{CL}$ and cross-entropy loss $\mathcal{L}_{CE}$. To explicitly disentangle speaker-specific features, EDGR employs a Gradient Reversal Layer (GR), which reverses the gradient of the speaker classification loss $\mathcal{L}_{ED}$ during backpropagation. To assess the impact of each component, we design the following five variants:


\begin{itemize}
\item {\textit{w/o $\mathcal{L}_{CE}$}}: This variant removes $\mathcal{L}_{CE}$ in Eq.~\eqref{eq:ID}. 
\item {\textit{w/o $\mathcal{L}_{CL}$}}: This variant removes $\mathcal{L}_{CL}$ in Eq.~\eqref{eq:ID}. 
\item {\textit{w/o $\mathcal{L}_{ID}$}}: This variant removes $\mathcal{L}_{ID}$ in Eq.~\eqref{eq:Total}, which is equivalent to removing the entire IDCFL module. 
\item {\textit{w/o ${GR}$}}: This variant removes ${GR}$ from EDGR. 
\item {\textit{w/o $\mathcal{L}_{ED}$}}: This variant removes $\mathcal{L}_{ED}$ from EDGR, which renders the GR layer ineffective and disables the entire EDGR module.
\end{itemize}



 The detailed results of these ablation studies are shown in Table~\ref{tab:loss}, leading to the following observations.
 

For the IDCFL module, \textit{w/o $L_{CL}$} results in lower performance than SIFLip, indicating that the contrastive learning in SIFLip improve the semantic consistency between paired visual and textual features by pulling positive pairs closer and pushing negative pairs apart, significantly improving the discrimination of visual features. The performance of \textit{w/o $\mathcal{L}_{CE}$} is even lower than that of \textit{w/o $\mathcal{L}_{CL}$}, indicating that the cross-entropy loss contributes to finer-grained visual-text alignment. \textit{w/o $\mathcal{L}_{ID}$} leads to substantial performance degradation on both seen and unseen speakers, highlighting the importance of learning visual features that are semantically aligned with text for enhancing overall performance and generalization.
\begin{table}[H]
\fontsize{10}{10}\selectfont
\caption{ Ablation study of joint loss function. }
\label{tab:loss}
    \centering
    \begin{tabular}{lcc}
        \toprule
      \multirow{2}{*}{\textbf{Methods}} & \multicolumn{2}{c}{\textbf{CMLR}} \\
        \cmidrule(lr){2-3}
         & \textbf{Seen (CER $\downarrow$)} & \textbf{Unseen (CER $\downarrow$)} \\
        \midrule
        \textbf{SIFLip} &\textbf{20.55} & \textbf{32.16}  \\
        \hline
        \textit{w/o $L_{CL}$} & 22.89 & 32.95  \\
       \textit{w/o $L_{CE}$} & 22.95 & 33.57  \\ 
       \textit{w/o $L_{ID}$} & 26.22 & 39.02  \\
       \hline
       
       \textit{w/o GR} & 20.96 & 33.07 \\
      \textit{w/o $L_{ED}$ }& 21.49 & 33.13 \\
     
        \bottomrule
    \end{tabular}
\end{table}

For the EDGR module, textit{w/o GR} exhibits a significant increase in CER for unseen speakers. This suggests that gradient inversion is effective in suppressing speaker-specific features in the backbone network by reversing the speaker classification loss in the case that the classifier is accurate. \textit{w/o $\mathcal{L}_{ED}$}, which leads to further performance degradation, confirming that the speaker sub-task can effectively filter out speaker-specific features.

\subsection{Visualization for IDCFL.} 
We delve deeper into visualizing the cross-modal alignment attained by the IDCFL module. Specifically, we pick one sample from the CMLR dataset to vividly illustrate the extent of alignment between visual and text features, as presented in Figure 5. In the heatmap, darker hues signify higher similarity and more optimal alignment, whereas lighter hues mirror lower similarity. These outcomes clearly demonstrate the efficacy of IDCFL in accomplishing cross-modal feature alignment.



\subsection{Hyperparameter Study}
In this subsection, we perform hyperparameter studies on CMLR dataset to evaluate the influence of $\alpha$ and $\beta$ in Eq.~\eqref{eq:Total} on model performance.

\noindent \textbf{Effect of the Hyper-parameter $\alpha$.} The weight for the Loss \( \mathcal{L}_{ID} \) is an important hyper-parameter, as it directly guides the backbone network to learn visual features aligned with the semantics of the spoken text. As shown in Figure~\ref{figure:line_bate}, the model achieves optimal performance at \( \alpha = 0.5 \).

\begin{figure}[ht]
\centering
\includegraphics[width=3.5in]{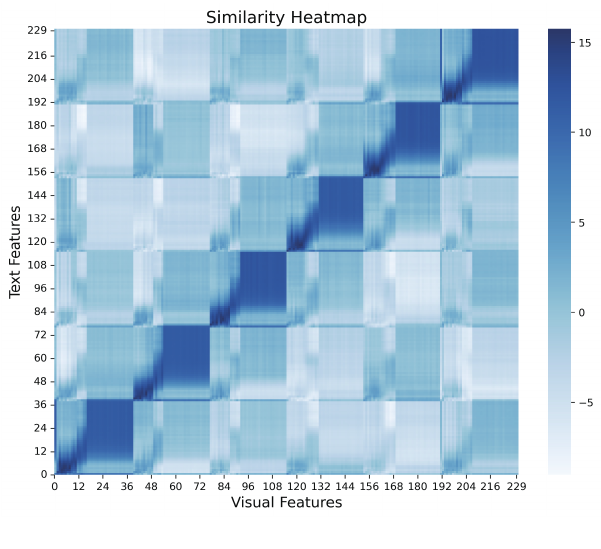}
\caption{Heatmap of Visual and Text Feature Similarity. The horizontal axis represents visual features and the vertical axis represents text features.}
\label{figure:heatmap}
\end{figure}

\begin{figure}[ht]
\centering
\includegraphics[width=3.5in]{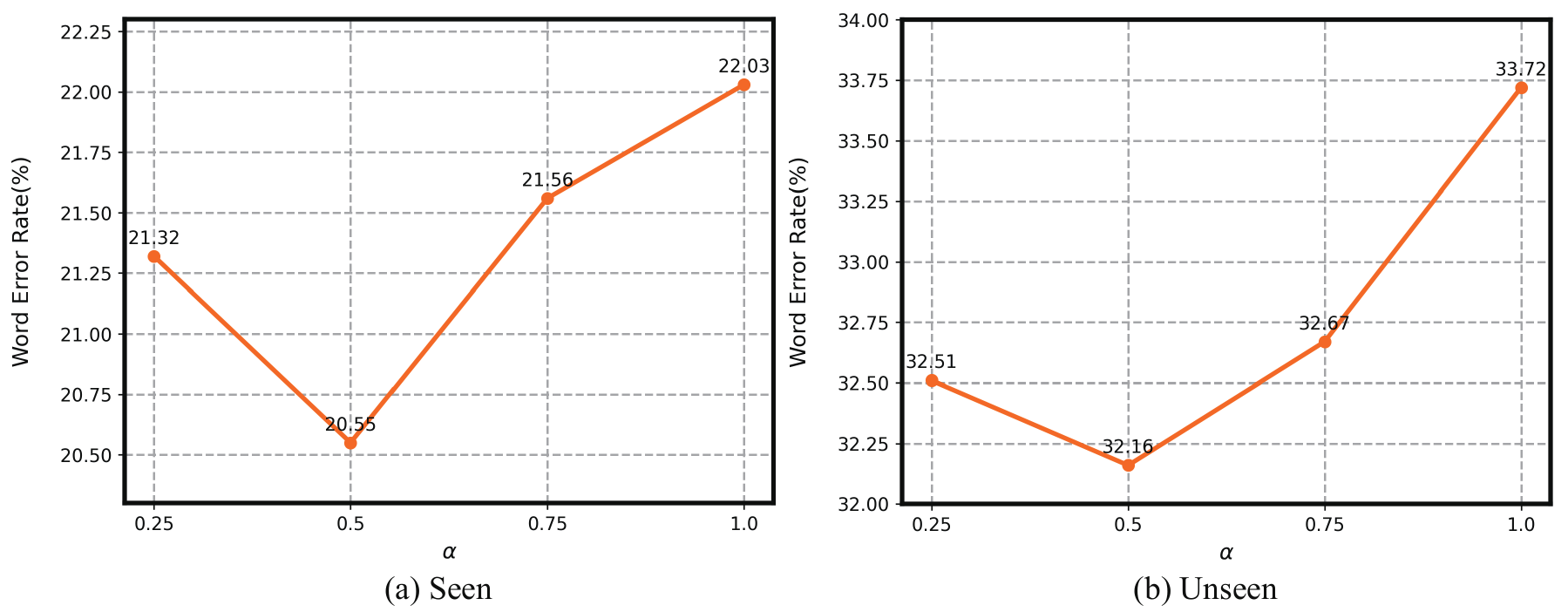}
\caption{ Effects of different hyper parameter $\alpha$ in Eq.~\eqref{eq:Total} on CMLR dataset. (a) Seen speakers. (b) Unseen speakers.}
\label{figure:line_gamma}
\end{figure}

\noindent \textbf{Effect of the Hyper-parameters $\beta$.} As shown in Figure~\ref{figure:line_bate}, the model achieves optimal generalization at \( \beta = 2.0 \). A moderate \( \beta \) effectively suppresses speaker-specific features. However, an overly large value may dominate optimization, hindering the extraction of speech-relevant visual features and degrading performance.
\begin{figure}[H]
\centering
\includegraphics[width=3.5in]{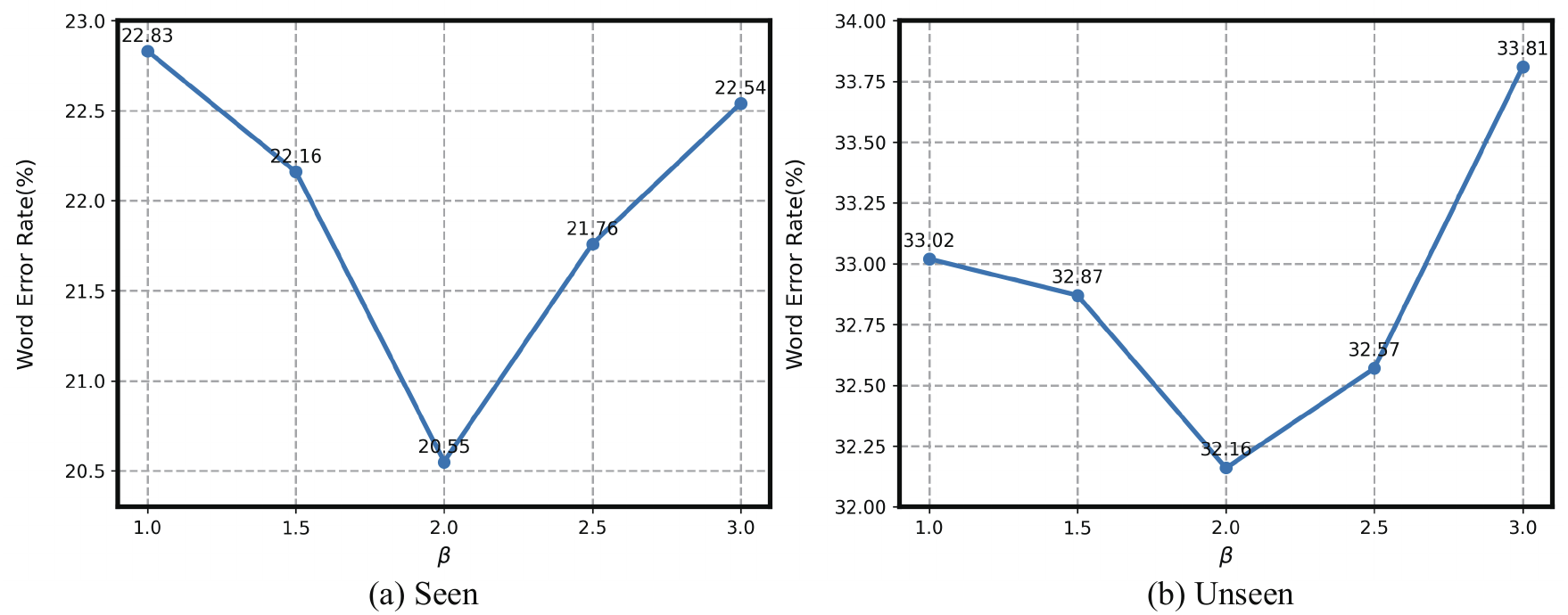}
\caption{ Effects of different hyper parameter $\beta$ in Eq.~\eqref{eq:Total} on CMLR dataset. (a) Seen speakers. (b) Unseen speakers.}
\label{figure:line_bate}
\end{figure}

\subsection{\textbf{Case Study}} 
\begin{figure}[hbt]
\centering
\includegraphics[width=\columnwidth]{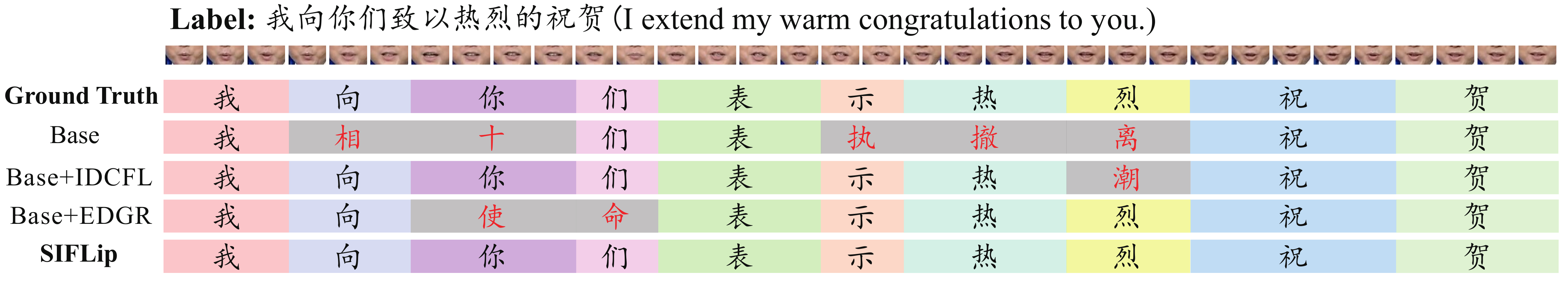}
\caption{Visualization of prediction examples on CMLR. Each colored block corresponds to its generated word. Gray blocks indicate prediction errors, and red font highlights word prediction errors.}
\label{fig:case study}
\end{figure}

To qualitatively analyze the effectiveness of the proposed model, we visualized one case from CMLR dataset, as shown in Figure \ref{fig:case study}. The results clearly indicate that the base predictions contain notable errors compared to the ground truth. With incorporates the IDCFL module, prediction accuracy improves significantly. Furthermore, incorporating the EDGR module further improves generalization to unseen speakers.

\section{\textbf{Conclusion}}\label{sec:con}
In this paper, we propose SIFLip, a novel lipreading framework designed to enhance model generalization by learning speaker-invariant visual features through dual-branch feature disentanglement. Our method disentangles speaker-specific features via two complementary mechanisms: (1) Implicit disentanglement, which aligns visual features with their corresponding text features to capture speaker-invariant features; and (2) Explicit disentanglement, which employs gradient reversal layers to adversarially suppress speaker-specific information during feature learning. Comprehensive evaluations on benchmark datasets demonstrate that SIFLip achieves state-of-the-art performance while significantly improving generalization to unseen speakers. In future work, we aim to extend our approach to side-view lipreading scenarios to further enhance its robustness and applicability in real-world situations. 




\bibliographystyle{ACM-Reference-Format}
\bibliography{sample-base}

\appendix

\end{document}